\def\year{2020}\relax
\documentclass[letterpaper]{article}
\usepackage{aaai20}
\usepackage{times}
\usepackage{helvet}
\usepackage{courier}

\usepackage{times}  %Required
\usepackage{helvet}  %Required
\usepackage{courier}  %Required
\usepackage{graphicx}  %Required

\usepackage{enumitem} % Customized lists
\setlist[itemize]{noitemsep} % Make itemize lists more compact
\usepackage{amsmath}
\usepackage{amsfonts}
\usepackage{amssymb}
\usepackage[mathscr]{euscript}
 \let\mathscr\relax
\usepackage[scr]{rsfso}

\usepackage{amsthm}

\usepackage{enumerate}
\usepackage{mathtools}
\usepackage{todonotes}
\usepackage{algorithm}
\usepackage[noend]{algpseudocode}
\usepackage{todonotes}
\usepackage{filecontents}
\usepackage{pgf}
\usepackage{pgfplots}
\usepackage{longtable}
\usepackage{array}
\usepackage{colortbl}
\usepackage{booktabs} % For \toprule, \midrule and \bottomrule
\usepackage{siunitx} % Formats the units and values
\usepackage{pgfplotstable} % Generates table from .csv
\usetikzlibrary{arrows,decorations,backgrounds,automata,shapes,patterns,decorations,backgrounds,plotmarks,calc,fit}
\usepackage{multirow}

\usepackage{float}

\newcommand{\idest}{{\it i.e.}}
\newcommand{\exemp}{{\it e.g.}}

\newcommand{\etal}{{\it et al.}}

\newtheorem{definition}{Definition}

\newcommand{\checkit}[1]{}

\newcommand{\prop}[1]{{\cal P}_{A}}

\usepackage{graphicx}
\title{The More the Merrier?! Evaluating the Effect of Landmark Extraction Algorithms on Landmark-Based Goal Recognition}

\author{
Kin Max Piamolini Gusmão, Ramon Fraga Pereira, and Felipe Meneguzzi \\
Pontifical Catholic University of Rio Grande do Sul (PUCRS), Brazil \\
\texttt{kin.gusmao@edu.pucrs.br}, \texttt{ramon.pereira@edu.pucrs.br}\\
\texttt{felipe.meneguzzi@pucrs.br}
}

\begin{document}

\maketitle

%----------------------------------------------------------------------------
\begin{abstract}
	
Recent approaches to goal and plan recognition using classical planning domains have achieved state of the art results in terms of both recognition time and accuracy by using heuristics based on planning landmarks. 
To achieve such fast recognition time these approaches use efficient, but incomplete, algorithms to extract only a subset of landmarks for planning domains and problems, at the cost of some accuracy. 
In this paper, we investigate the impact and effect of using various landmark extraction algorithms capable of extracting a larger proportion of the landmarks for each given planning problem, up to exhaustive landmark extraction. 
We perform an extensive empirical evaluation of various landmark-based heuristics when using different percentages of the full set of landmarks. 
Results show that having more landmarks does not necessarily mean achieving higher accuracy and lower spread, as the additional extracted landmarks may not necessarily increase be helpful towards the goal recognition task.
\end{abstract}
%----------------------------------------------------------------------------

%----------------------------------------------------------------------------
\section{Introduction}

Anticipating and recognizing correctly the intended goal that an observed agent aims to achieve based on its interactions in an environment is an important task for several real-world applications~\cite{Oh2014}, such as intent recognition for elder-care~\cite{ProblemsWithElderCare_AAAI2002}, exploratory domain models~\cite{PR_EXP_Mirsky2017,Oh2013}, offline and online goal recognition in latent space~\cite{GR_LatentSpace_Amado2018,LSTM_GR,Amado2019_DEMO}, and others. 
Most approaches to goal and plan recognition rely on either plan libraries~\cite{AvrahamiZilberbrandK_IJCAI2005,PRGeib_AI_2009,PR_Application_Amir_2013,Mirsky2017plan} or planning domain theory~\cite{RamirezG_IJCAI2009,RamirezG_AAAI2010,PattisonGoalRecognition_2010,GoalRecognitionDesign_Keren2014,Sohrabi_IJCAI2016,Masters_IJCAI2017}. 
Recent work on goal recognition as planning has avoided running a full-fledged planner for recognizing goals, and recent approaches in the literature have successfully exploited the use of well-known automated planning techniques, such as planning graphs~\cite{NASA_GoalRecognition_IJCAI2015} and landmarks~\cite{RamonNirMeneguzzi_AAAI2017,PereiraOrenMeneguzzi_PAIR2017,PereiraAAMAS_2017}. 
Thus, as a result of exploiting planning techniques, such approaches have shown that it is possible to recognize goals and plans not only accurately, but also very quickly.

In this paper, we investigate the effect of using various landmark extraction algorithms over the landmark-based heuristic to goal recognition proposed by Pereira, Oren, and Meneguzzi~\shortcite{RamonNirMeneguzzi_AAAI2017}. 
For extracting landmarks, we use five landmark extraction algorithms~\cite{zhugivan,Hoffmann2004_OrderedLandmarks,rhw,hm} from the planning literature. 
To do so, we use an exhaustive extraction algorithm (i.e., an extraction approach that exhaustively checks if all facts are landmark by using a relaxed planning graph), and use other extraction algorithms that extract only a subset of landmarks~\cite{zhugivan,Hoffmann2004_OrderedLandmarks,rhw,hm}. 
Thus, the main contribution of this paper is investigating the real impact of using more or fewer landmarks in the landmark-based goal recognition heuristics.

We conduct extensive experiments to empirically evaluate the impact and effect of using a variety different landmark extraction algorithms over landmark-based recognition heuristics using well-known recognition datasets~\cite{Pereira2017_dataset} with missing and full observations, and noisy, missing, and full observations. 
Results show that using more landmarks does not necessarily lead to improved precision and accuracy of the landmark-based heuristics, as the quality of the extracted landmarks is generally more important than the quantity.

The remainder of this paper is organized as follows. 
Section~\ref{section:background} provides essential background on planning, goal recognition, and landmarks. 
We review the landmark-based heuristic approaches we use along with various landmark extraction algorithms in Section~\ref{section:landmark_goalrecognition}. 
In Section~\ref{section:experiments_evaluation}, we proceed to evaluate empirically the recognition heuristics we review.
Finally, in Section~\ref{section:conclusions}, we conclude this paper by discussing the real impact of using more or fewer landmarks in the heuristics, and provide future directions of how such heuristics could be improved by taking advantage of more landmarks. 

%----------------------------------------------------------------------------
\section{Background}\label{section:background}

%########################################################################
\subsection{Planning}

Planning is the problem of finding a sequence of actions (\idest, a plan) that achieves a goal from an initial state~\cite{AutomatedPlanning_Book2011}.
A \textit{state} is a finite set of facts that represent logical values according to some interpretation. 
\textit{Facts} can be either positive, or negated ground predicates.
A predicate is denoted by an n-ary predicate symbol $p$ applied to a sequence of zero or more terms ($\tau_1$, $\tau_2$, ..., $\tau_n$). 
An \textit{operator} is represented by a triple $a$ $=$ $\langle$\textit{name}($a$), \textit{pre}($a$), \textit{eff}($a$)$\rangle$ where \textit{name}($a$) represents the description or signature of $a$; \textit{pre}($a$) describes the preconditions of $a$ --- a set of facts or predicates that must exist in the current state for $a$ to be executed; $\mathit{eff}($a$)=\mathit{eff}(a)^+ \cup \mathit{eff}(a)^-$ represents the effects of $a$, with \textit{eff}($a$)$^+$  an \textit{add-list} of positive facts or predicates, and \textit{eff}($a$)$^-$ a \textit{delete-list} of negative facts or predicates. 
When we instantiate an operator over its free variables, we call the resulting ground operator an \emph{action}. 
A \textit{planning instance} is represented by a triple $\Pi = \langle\Xi, \mathcal{I}, G\rangle$, in which  $\Xi = \langle\Sigma,\mathcal{A}\rangle$ is a \textit{planning domain definition}; $\Sigma$ consists of a finite set of facts and $\mathcal{A}$ a finite set of actions; $\mathcal{I}$ $\subseteq$ $\Sigma$ is the initial state; and $G$ $\subseteq$ $\Sigma$ is the goal state. 
A \textit{plan} is a sequence of actions $\pi = \langle a_1, a_2, ..., a_n \rangle$ that modifies the initial state $\mathcal{I}$ into one in which the goal state $G$ holds by the successive execution of actions in a plan $\pi$. While actions have an associated cost, as in classical planning, in this paper we assume that this cost is 1 for all actions. A plan $\pi$ is considered optimal if its cost, and thus length, is minimal.

%########################################################################
\subsection{Goal Recognition}

Goal recognition is the task of discerning the intended goal of autonomous agents or humans by observing their interactions in a particular environment~\cite[Chapter 1]{ActivityIntentPlanRecogition_Book2014}. 
Such observed interactions are defined as available evidence that can be used to recognize goals. We formally define the problem of goal recognition over planning domain theory by adopting the formalism proposed by Ram{\'{\i}}rez and Geffner~\shortcite{RamirezG_IJCAI2009,RamirezG_AAAI2010}, as follows in Definition~\ref{def:planRecognition}. 

\begin{definition}[\textbf{Goal Recognition Problem}]\label{def:planRecognition}
A goal recognition problem is a tuple $T_{GR}$ $=$ $\langle\Xi,\mathcal{I} ,\mathcal{G}, O\rangle$, in which $\Xi = \langle\Sigma, \mathcal{A}\rangle$ is a planning domain definition; $\mathcal{I}$ is the initial state; $\mathcal{G}$ is the set of possible goals, which include the correct intended goal $G^{*}$ (\idest, $G^{*}$ $\in$ $\mathcal{G}$); and $O$ $=$ $\langle$$o_1$, $o_2$, ..., $o_n$$\rangle$ is an observation sequence of executed actions, with each observation $o_i \in \mathcal{A}$.
\end{definition}

The ideal solution for a goal recognition problem is finding the correct intended goal $G^{*} \in \mathcal{G}$ that the observation sequence $O$ of a plan execution achieves. 
An observation sequence can be full or partial --- in a full observation sequence we observe all actions of an agent's plan; in a partial observation sequence, only a sub-sequence of actions are observed. A noisy observation sequence contains one or more actions (or a set of facts) that might not be part of a plan that achieves a particular goal, \exemp, when a sensor fails and generates abnormal or spurious readings.

%########################################################################
\subsection{Landmarks}

In the planning literature, landmarks are defined as necessary fact (or actions) that must be true (or executed) at some point along all valid plans that achieve a particular goal from an initial state. Landmarks are often partially ordered based on the sequence in which they must be achieved. 
Hoffman \etal~\shortcite{Hoffmann2004_OrderedLandmarks} define fact landmarks as follows:

\begin{definition}[\textbf{Fact Landmark}]\label{def:planLandmark}
Given a planning instance $\Pi = \langle \Xi, \mathcal{I}, G\rangle$, a formula $L$ is a fact landmark in $\Pi$ iff $L$ is true at some point along all valid plans that achieve $G$ from $\mathcal{I}$. 
A landmark is a type of formula (\exemp, a conjunctive or disjunctive formula) over a set of facts that must be satisfied at some point along all valid plan executions.
\end{definition}

Hoffman \etal~\shortcite{Hoffmann2004_OrderedLandmarks} proves that the process of generating  all landmarks and deciding their ordering is PSPACE-complete, which is exactly the same complexity as deciding plan existence~\cite{PlanningComplexity_Bylander1994}. 
Thus, to operate efficiently, most landmark extraction algorithms extract only a subset of landmarks for a given planning instance.

%########################################################################
\subsection{Landmark Extraction Algorithms}

% We use various landmark extraction algorithms to investigate whether more or fewer landmarks impact in the recognition accuracy of landmark-based heuristics for goal recognition, and now, we present in detail the landmark extraction algorithms we use in this paper.
In this paper, we use the following landmark extraction algorithms to investigate how the number of landmarks impacts on the recognition accuracy of landmark-based heuristics for goal recognition.

\noindent \textit{Exhaust}: The first algorithm is an exhaustive extraction approach, its name says for itself, and we denote this algorithm as \textit{Exhaust}. This algorithm exhaustively extracts landmarks for a given planning instance. Namely, this algorithm uses a Relaxed Planning Graph (RPG) and exhaustively checks every fact in the RPG for if it is a landmark or not. This is done by removing the fact from the RPG and checking if the goal is still reachable without the given fact, and if not, such fact is considered as a landmark. The number of landmarks extracted by this algorithm is used as a baseline in our experiments, as it can extract all landmarks for a planning instance.

\noindent $h^m$: \citeauthor{hm}~\shortcite{hm} developed a landmark extraction algorithm that performs a transformation of the original problem $\Pi$, originating a new problem $\Pi^m$, in which each fact is a set of facts of size $m$, originated from the original problem's facts. The actions are obtained by adding facts that are not required or caused by any action but might be true during plan development, to the action's preconditions and effects. The result is a problem without delete effects that yet has information on the delete effects of the original problem, hence allowing the extraction of landmarks that take delete effects into count. This extraction algorithm is denoted as $h^m$.

\noindent \textit{RHW}: In~\cite{rhw}, \citeauthor{rhw}~\shortcite{rhw} develop a landmark extraction algorithm that starts the process by selecting an initial fact landmark, and from this initial landmark, it creates disjunctive sets from the preconditions of the actions that are first achievers of the initial landmark. Each disjunctive set is then recorded as a landmark, and ordered before the initial landmark. This extraction process is then repeated for all recorded landmarks. We denote this algorithm as \textit{RHW}.

\noindent \textit{Zhu \& Givan}: Zhu and Givan~\shortcite{zhugivan} developed a landmark extraction algorithm that works differently than the ones mentioned above. This algorithm works by propagating labels across the planning graph, where each label is a fact or an action. A fact or action at a level $i$ must be labeled with any fact or action that must occur in any $i$-step plan that reaches it. It starts by labeling each action in the first action level with itself. Every subsequent action level is then labeled with the union of the labels on its precondition fact nodes, while every subsequent fact node is labeled with the intersection of the labels on the action nodes that reach it. At the last level, every label on a goal node is considered a landmark. This algorithm is denoted as \textit{Zhu \& Givan}.

\noindent \textit{Hoffmann et al.}: The extraction algorithm originally used by \cite{RamonNirMeneguzzi_AAAI2017} is the landmark extraction algorithm of Hoffman \etal~\shortcite{Hoffmann2004_OrderedLandmarks}. 
Initially, this algorithm builds an RPG (ignoring all delete effects of all actions) from the initial state to the goal state, and starts selecting all facts in goal state as candidate landmarks. 
Afterward, it selects the preconditions for all actions that achieve each candidate landmark, checking if those are landmarks by removing them from the graph and checking the reachability of the goal. 
After, it records as landmarks all preconditions that passed this check and then repeats the process for every fact level on the graph back to the initial state. 
Similar to the \textit{Exhaust} method, this algorithm evaluates whether a candidate landmark is indeed a landmark by testing the solvability of the problem by removing all actions that achieve such candidate landmark, and if the problem is unsolvable, then this candidate landmark is indeed a landmark. 
We denote this algorithm as \textit{Hoffmann et al.}

%----------------------------------------------------------------------------
\section{Landmark-Based Goal Recognition}\label{section:landmark_goalrecognition}

We now describe the goal recognition heuristics that rely on planning landmarks that we use to evaluate the effect of using different landmark extraction algorithms. Such heuristics have proved to be accurate and very quick for recognizing goals over a variety of domain models~\cite{PereiraMeneguzzi_ECAI2016,RamonNirMeneguzzi_AAAI2017}. 

The first landmark-based heuristic proposed by Pereira, Oren, and Meneguzzi~\shortcite{RamonNirMeneguzzi_AAAI2017} is called \emph{goal completion heuristic}, and denoted as $\mathit{h}_{gc}$. 
Basically, this heuristic computes a score for a goal $G$ by calculating the ratio between the number achieved landmarks for $G$ and the total number of extracted landmarks for $G$. 
This score represents the percentage of completion of goal based on the ratio of achieved landmarks and the total number of landmarks.

As an extension of $\mathit{h}_{gc}$, the second heuristic developed by Pereira, Oren, and Meneguzzi~\shortcite{RamonNirMeneguzzi_AAAI2017} exploits the concept of \emph{landmark uniqueness value}~\cite{RamonNirMeneguzzi_AAAI2017}, which is a value that represents how unique a landmark is among the set of landmarks for all possible goals. 
This heuristic is called \emph{landmark uniqueness heuristic}, and denoted as $\mathit{h}_{uniq}$. 
Thus, by using this uniqueness value, $\mathit{h}_{uniq}$ estimates which possible goal is most likely the intended one by summing the uniqueness values of the landmarks achieved in the observations. 

% %########################################################################
% \subsection{Goal Completion Heuristic}
%
% %########################################################################
% \subsection{Uniqueness Heuristic}

%----------------------------------------------------------------------------
\section{Experiments and Evaluation}\label{section:experiments_evaluation}

In this section, we present the experiments and evaluations we carried out from using various extraction algorithms over the landmark-based goal recognition heuristics.

%########################################################################
\subsection{Domains and Setup}

For evaluating each one of the landmark extraction algorithms using both recognition heuristics, we executed several tests using datasets created by Pereira and Meneguzzi \shortcite{Pereira2017_dataset}, containing several non-trivial recognition problems. 
These datasets contain goal recognition problems from 15 classical planning domains and include problems with noisy observations. 
The domains we used are: Blocks World, Campus, Depots, Dock Worker Robots, Driverlog, Easy IPC Grid, Ferry, Intrusion Detection, Logistics, Miconic, Rovers, Satellite, Sokoban and Zeno Travel. 
The Kitchen domain has been removed from our evaluation, as it is an adaptation of an HTN planning domain and it caused some issues when using some of the landmark extractors.

Each domain in these datasets includes recognition problems with partial and full observations. 
Partial observations vary the level (percentage) of observability between 10\%, 30\%, 50\% and 70\% of actions observed for missing observations, and 100\% for full observations. 
For problems with noisy observations, the level (percentage) of observability varies between 25\%, 50\% and 75\% of observed actions for missing observations, and consequently 100\% for full observations.

% Fro

% With this experimentation setup, we aim to have a complete vision on how each of the heuristics behave when allied to each of the algorithms, with various recognition thresholds.

%########################################################################
\subsection{Evaluation Metrics}

To evaluate the recognition heuristics, we use three metrics: recognition time (Time), accuracy (Acc\%) and Spread in $\mathcal{G}$ (S in $\mathcal{G}$). 
The recognition time metric is simply the time in seconds that the algorithm took to return the set of recognized goals, including the time for extracting the landmarks. 
Accuracy is a percentage that represents the average number of problems in which the correct goal was among the recognized goals list. 
Finally, Spread in $\mathcal{G}$ is the average number of returned goals, when multiple goal hypotheses were tied in the recognition algorithm. 
To have a concise precision metric of the approach, we combine accuracy and Spread in $\mathcal{G}$ to obtain a third metric. 
This metric can be considered as a precision metric and is obtained by calculating the ratio between accuracy and Spread in $\mathcal{G}$.

Since our goal is to find out if there is a relation between the number of extracted landmarks and the effectiveness of a landmark-based goal recognition technique, we also use a metric to evaluate the extraction capability of each landmark extraction algorithm. 
We do this by calculating the ratio between the number of landmarks extracted by each algorithm and the number of landmarks extracted by the \textit{Exhaust} algorithm, since it can extract all landmarks in the planning instance. The result is the percentage of extracted landmarks.

%########################################################################
\subsection{Results: Missing and Full Observations}

We now present the results for datasets with missing and full observations.
Table \ref{tab:results} shows the results comparing the use of the five different extraction algorithms along with the landmark-based heuristics. 
We can see the average number of landmarks extracted, represented by $\mathcal{L}$, average recognition time in seconds, average accuracy (Acc\%) and average Spread in $\mathcal{G}$ (S in $\mathcal{G}$) for each combination of extraction algorithm and threshold used for heuristics $h_{gc}$ and $h_{uniq}$. 
Columns represent different levels of observability.

\begin{table*}[ht!]
\centering
\fontsize{6.5}{11}\selectfont
\setlength\tabcolsep{3pt}
\begin{tabular}{rrrrr@{\hspace*{8mm}}rrr@{\hspace*{8mm}}rrr@{\hspace*{8mm}}rrr@{\hspace*{8mm}}rrl}
\toprule	
\hline

&
& \multicolumn{3}{c}{\bf 10\%}                                                             
& \multicolumn{3}{c}{\bf 30\%}                                                             
& \multicolumn{3}{c}{\bf 50\%}                                                             
& \multicolumn{3}{c}{\bf 70\%}                                                             
& \multicolumn{3}{c}{\bf 100\%}                                                          \\ \hline

\textbf{Approach}
& $|\mathcal{L}|$
& \multicolumn{1}{r}{\bf Time} & \multicolumn{1}{r}{\bf Acc \%} & \multicolumn{1}{l}{\bf S in $\mathcal{G}$} 
& \multicolumn{1}{r}{\bf Time} & \multicolumn{1}{r}{\bf Acc \%} & \multicolumn{1}{l}{\bf S in $\mathcal{G}$} 
& \multicolumn{1}{r}{\bf Time} & \multicolumn{1}{r}{\bf Acc \%} & \multicolumn{1}{l}{\bf S in $\mathcal{G}$} 
& \multicolumn{1}{r}{\bf Time} & \multicolumn{1}{r}{\bf Acc \%} & \multicolumn{1}{l}{\bf S in $\mathcal{G}$} 
& \multicolumn{1}{r}{\bf Time} & \multicolumn{1}{r}{\bf Acc \%} & \multicolumn{1}{r}{\bf S in $\mathcal{G}$} \\ \hline
      $\mathit{h_{gc}}$ (Exhaust $\theta = 0$) 
        & 36.9 & 5.848 & 63.4\% & 1.598
             & 5.565 & 84.2\% & 1.259
             & 6.708 & 89.9\% & 1.114
             & 6.403 & 96.4\% & 1.048
             & 6.874 & 99.6\% & 1.025
            \\
        $\mathit{h_{gc}}$ (Exhaust $\theta = 10$) 
        & 36.9 & 5.855 & 88.7\% & 3.378
             & 5.558 & 96.9\% & 2.740
             & 6.724 & 98.9\% & 2.421
             & 6.377 & 99.6\% & 2.170
             & 6.894 & 100.0\% & 1.869
            \\
        $\mathit{h_{gc}}$ ($h^m$ $\theta = 0$) 
        & 20.6 & 19.575 & 66.7\% & 1.634
             & 19.844 & 83.2\% & 1.249
             & 23.836 & 89.7\% & 1.143
             & 21.725 & 96.5\% & 1.054
             & 24.013 & 99.7\% & 1.046
            \\
        $\mathit{h_{gc}}$ ($h^m$ $\theta = 10$) 
        & 20.6 & 19.540 & 83.6\% & 2.819
             & 19.860 & 92.8\% & 2.177
             & 23.774 & 97.1\% & 1.930
             & 21.677 & 99.2\% & 1.736
             & 24.220 & 100.0\% & 1.474
            \\
        $\mathit{h_{gc}}$ (RHW $\theta = 0$) 
        & 23.5 & 5.793 & 64.8\% & 1.637
             & 5.521 & 81.6\% & 1.251
             & 6.664 & 89.1\% & 1.137
             & 6.342 & 96.3\% & 1.062
             & 6.872 & 99.5\% & 1.051
            \\
        $\mathit{h_{gc}}$ (RHW $\theta = 10$) 
        & 23.5 & 5.785 & 80.0\% & 2.735
             & 5.536 & 91.2\% & 2.215
             & 6.650 & 96.3\% & 2.000
             & 6.328 & 98.6\% & 1.787
             & 6.870 & 100.0\% & 1.509
            \\
        $\mathit{h_{gc}}$ (Zhu \& Givan $\theta = 0$) 
        & 19.9 & 5.798 & 66.4\% & 1.657
             & 5.523 & 83.1\% & 1.262
             & 6.683 & 89.7\% & 1.147
             & 6.338 & 96.4\% & 1.060
             & 6.871 & 99.7\% & 1.054
            \\
        $\mathit{h_{gc}}$ (Zhu \& Givan $\theta = 10$) 
        & 19.9 & 5.812 & 81.9\% & 2.768
             & 5.534 & 92.6\% & 2.213
             & 6.679 & 96.5\% & 1.947
             & 6.331 & 98.6\% & 1.747
             & 6.886 & 100.0\% & 1.483
            \\
        $\mathit{h_{gc}}$ (Hoffmann $\theta = 0$) 
        & 18.8 & 11.283 & 61.3\% & 1.630
             & 10.648 & 77.1\% & 1.268
             & 13.259 & 86.1\% & 1.149
             & 12.500 & 94.1\% & 1.072
             & 13.691 & 99.5\% & 1.056
            \\
        $\mathit{h_{gc}}$ (Hoffmann $\theta = 10$) 
        & 18.8 & 11.259 & 77.8\% & 2.743
             & 10.721 & 87.4\% & 2.141
             & 13.280 & 92.5\% & 1.850
             & 12.509 & 96.5\% & 1.679
             & 13.702 & 100.0\% & 1.454
            \\
        \hline
$\mathit{h_{uniq}}$ (Exhaust $\theta = 0$) 
        & 36.9 & 6.094 & 56.7\% & 1.153
             & 5.681 & 76.8\% & 1.070
             & 6.366 & 84.7\% & 1.035
             & 5.938 & 93.4\% & 1.022
             & 6.874 & 99.2\% & 1.025
            \\
        $\mathit{h_{uniq}}$ (Exhaust $\theta = 10$) 
        & 36.9 & 6.086 & 71.3\% & 1.881
             & 5.696 & 87.1\% & 1.537
             & 6.358 & 91.0\% & 1.325
             & 5.909 & 97.2\% & 1.244
             & 6.895 & 100.0\% & 1.130
            \\
        $\mathit{h_{uniq}}$ ($h^m$ $\theta = 0$) 
        & 20.6 & 20.892 & 58.0\% & 1.218
             & 19.494 & 76.0\% & 1.071
             & 22.266 & 85.3\% & 1.040
             & 20.448 & 94.2\% & 1.028
             & 23.853 & 99.7\% & 1.046
            \\
        $\mathit{h_{uniq}}$ ($h^m$ $\theta = 10$) 
        & 20.6 & 20.880 & 69.7\% & 1.772
             & 19.467 & 84.6\% & 1.429
             & 22.237 & 90.9\% & 1.287
             & 20.408 & 97.1\% & 1.203
             & 24.033 & 100.0\% & 1.112
            \\
        $\mathit{h_{uniq}}$ (RHW $\theta = 0$) 
        & 23.5 & 6.033 & 56.4\% & 1.214
             & 5.665 & 74.8\% & 1.067
             & 6.313 & 85.1\% & 1.039
             & 5.854 & 93.8\% & 1.029
             & 6.802 & 99.5\% & 1.051
            \\
        $\mathit{h_{uniq}}$ (RHW $\theta = 10$) 
        & 23.5 & 6.025 & 69.3\% & 1.815
             & 5.639 & 84.1\% & 1.453
             & 6.301 & 90.4\% & 1.306
             & 5.871 & 96.7\% & 1.216
             & 6.774 & 100.0\% & 1.121
            \\
        $\mathit{h_{uniq}}$ (Zhu \& Givan $\theta = 0$) 
        & 19.9 & 6.031 & 56.8\% & 1.212
             & 5.668 & 75.7\% & 1.070
             & 6.315 & 85.0\% & 1.042
             & 5.891 & 93.9\% & 1.031
             & 6.785 & 99.7\% & 1.054
            \\
        $\mathit{h_{uniq}}$ (Zhu \& Givan $\theta = 10$) 
        & 19.9 & 6.028 & 69.3\% & 1.788
             & 5.647 & 84.9\% & 1.441
             & 6.284 & 90.7\% & 1.291
             & 5.882 & 96.8\% & 1.201
             & 6.836 & 100.0\% & 1.112
            \\
        $\mathit{h_{uniq}}$ (Hoffmann $\theta = 0$) 
        & 18.8 & 11.903 & 53.4\% & 1.308
             & 11.076 & 70.8\% & 1.117
             & 12.383 & 80.2\% & 1.039
             & 11.460 & 90.8\% & 1.032
             & 13.610 & 98.5\% & 1.043
            \\
        $\mathit{h_{uniq}}$ (Hoffmann $\theta = 10$) 
        & 18.8 & 11.915 & 65.3\% & 1.868
             & 11.044 & 79.9\% & 1.486
             & 12.379 & 86.9\% & 1.328
             & 11.422 & 93.9\% & 1.236
             & 13.715 & 99.2\% & 1.120
            \\
        \hline
\bottomrule
\end{tabular}
\caption{Experiments and evaluations with missing and full observations.}
\label{tab:results}	
\end{table*}

We can see that even with 100\% of actions being observed, the heuristic recognition algorithms do not yield 100\% accuracy. 
There are some cases, for instance, in Driverlog and Logistics for $\mathit{h_{gc}}$, in which the real goal had more total landmarks than a wrong candidate goal, but only a few extra achieved landmarks than the wrong one. 
As a result, the heuristic chooses the wrong goal instead the correct one, especially with lower threshold values. 

We can also see that the extraction $h^m$  algorithm has the highest recognition time in comparison to all algorithms. 
\textit{Hoffmann et al.} has the second highest recognition time, while other algorithms come in third with similar recognition time.

\begin{figure}[h!]
    \centering
    \includegraphics[width=0.45\textwidth]{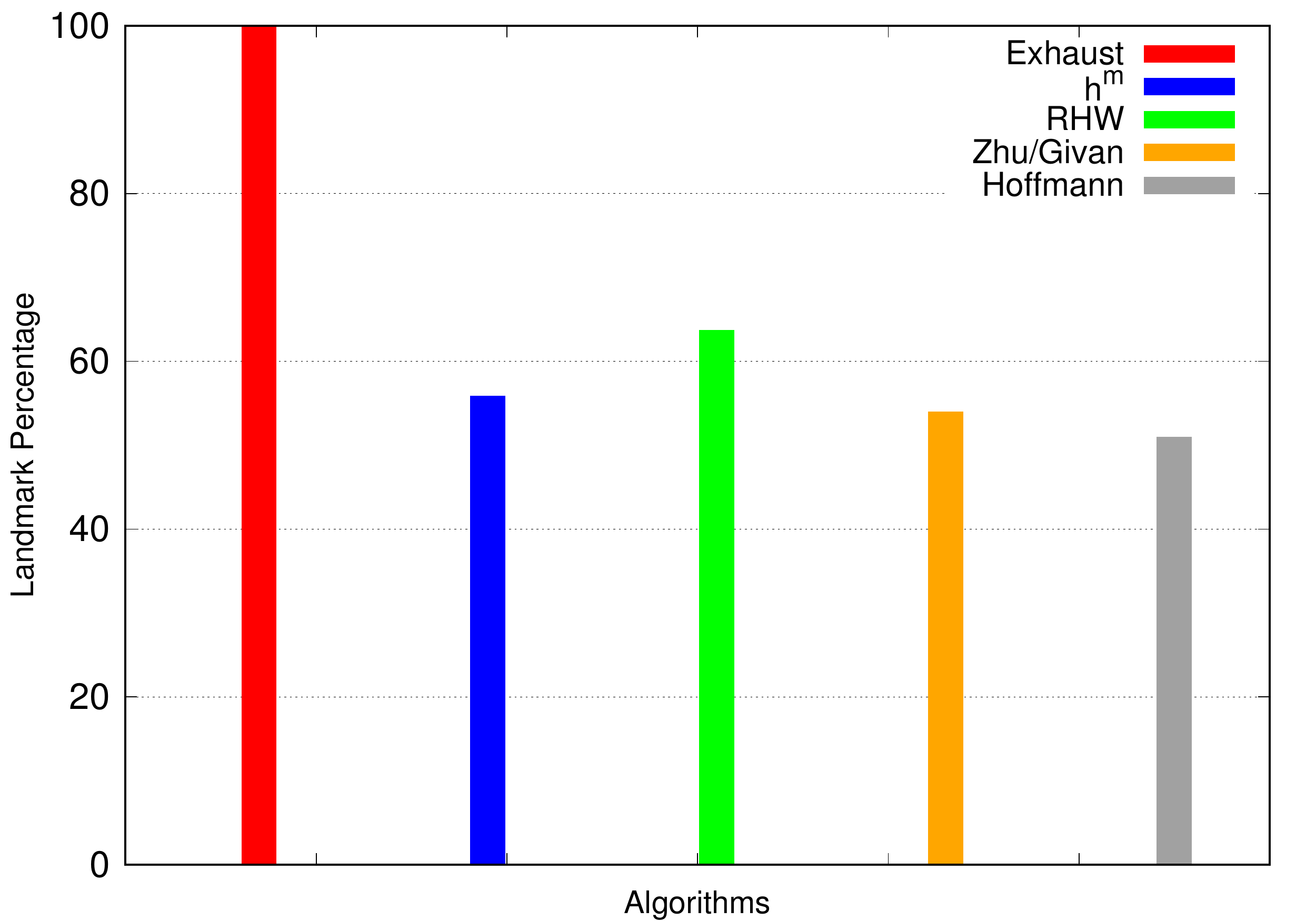}
    \caption{Percentage of extracted landmarks by algorithm with missing and full observations.}
    \label{fig:percentage}
\end{figure}

Figure~\ref{fig:percentage} shows the average percentage of extracted landmarks by each extraction algorithm we used in our experiments for Table~\ref{tab:results}. 
Note that, after \textit{Exhaust}, \textit{RHW} was the extraction algorithm that managed to extract the highest number of landmarks, on average, followed by $h^m$, \textit{Zhu \& Givan}, and finally \textit{Hoffman et al.}.

\begin{figure}[h!]
\centering
\includegraphics[width=0.45\textwidth]{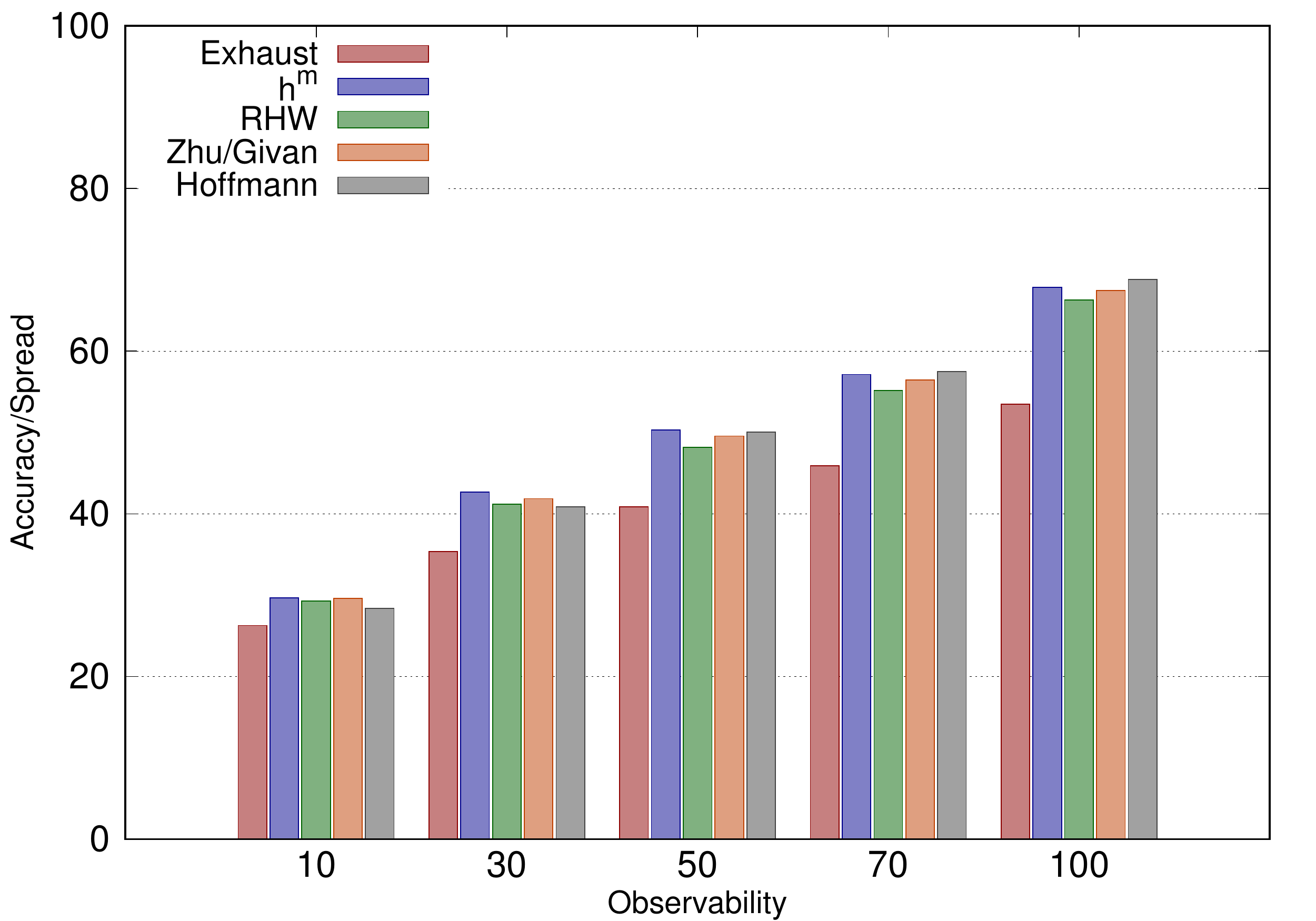}
\caption{Accuracy/Spread in $\mathcal{G}$ ratio for $\mathit{h_{gc}}$ with missing and full observations.}
\label{fig:accuracy_gc}
\end{figure}

\begin{figure}[h!]
\centering
\includegraphics[width=0.45\textwidth]{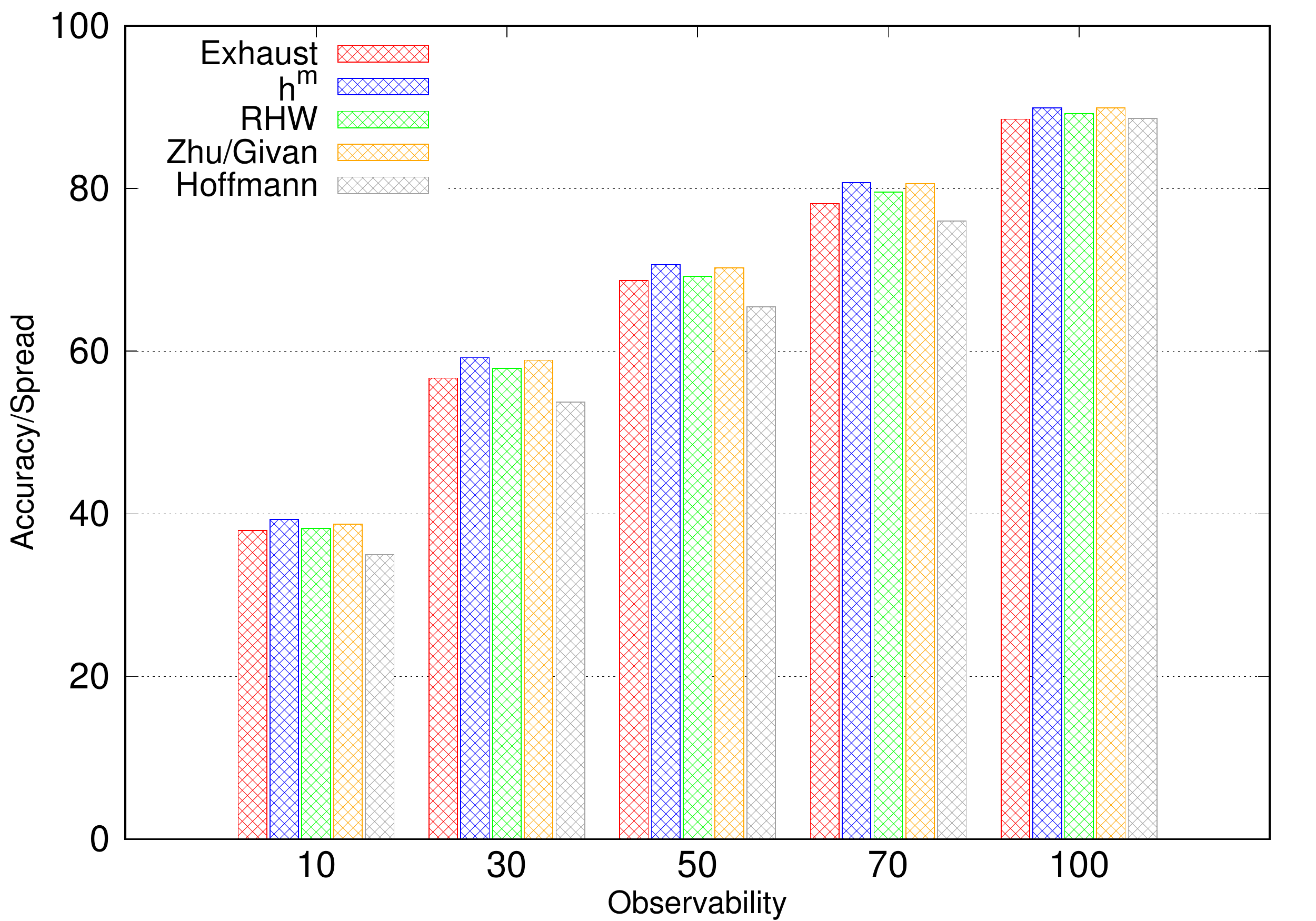}
\caption{Accuracy/Spread in $\mathcal{G}$ ratio for $\mathit{h_{uniq}}$ with missing and full observations.}
\label{fig:accuracy_uniq}
\end{figure}

Figures \ref{fig:accuracy_gc} and \ref{fig:accuracy_uniq} show the average Accuracy/Spread in $\mathcal{G}$ ratio with a threshold $\theta$ value of 10 for each combination of heuristic, extraction algorithm, and the level of observability. 
Although \textit{Exhaust} and \textit{RHW} managed to extract the highest number of landmarks, $h^m$ was the algorithm that led both heuristics to the highest Accuracy/Spread in $\mathcal{G}$ ratio, leaving even \textit{Exhaust} behind.

Based on the results of Figures~\ref{fig:accuracy_gc} and~\ref{fig:accuracy_uniq}, we can see that that the amount of extracted landmarks is not the only factor that affects the effectiveness for recognition using landmarks. 
We note that the quality of the extracted landmarks and how well they inform the heuristics cause real impact in the recognition process. 
We believe this is the reason $\mathit{h_{uniq}}$ yields a higher Accuracy/Spread ratio in the datasets with missing and full observations when compared to $\mathit{h_{gc}}$. 
The $\mathit{h_{uniq}}$ heuristic considers the degree of information provided by a landmark (\idest, \emph{landmark uniqueness values}), instead of just estimating using the amount of landmarks, as $\mathit{h_{gc}}$ does. 
The $\mathit{h_{uniq}}$ heuristic can filter relatively uninformative landmarks, assigning a greater \emph{landmark uniqueness value} for those that are found in fewer goals, hence better informing the heuristic.

\begin{figure}[h!]
\centering
\includegraphics[width=0.45\textwidth]{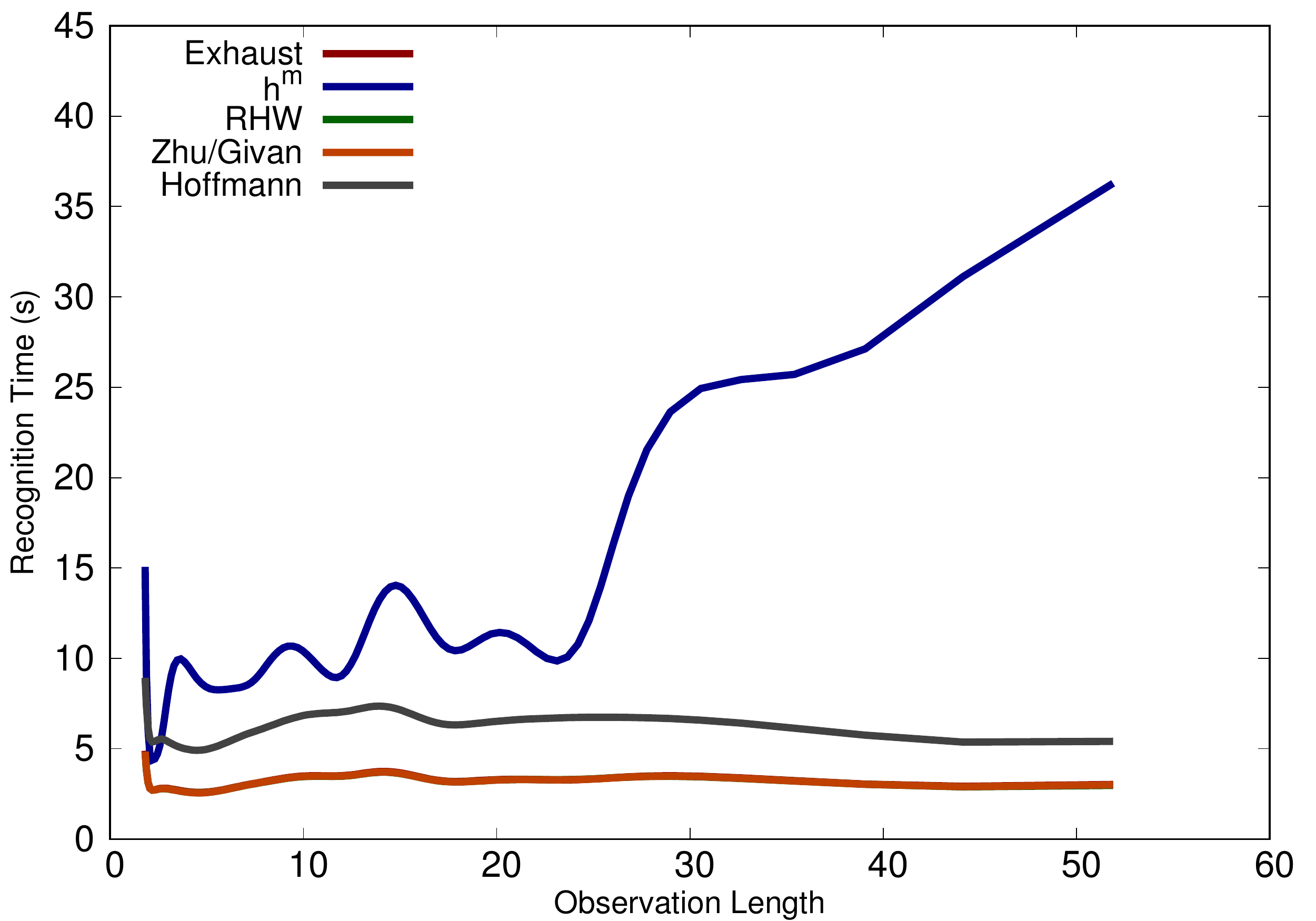}
\caption{Recognition time for $\mathit{h_{gc}}$ with missing and full observations.}
\label{fig:time_gc}
\end{figure}

\begin{figure}[h!]
\centering
\includegraphics[width=0.45\textwidth]{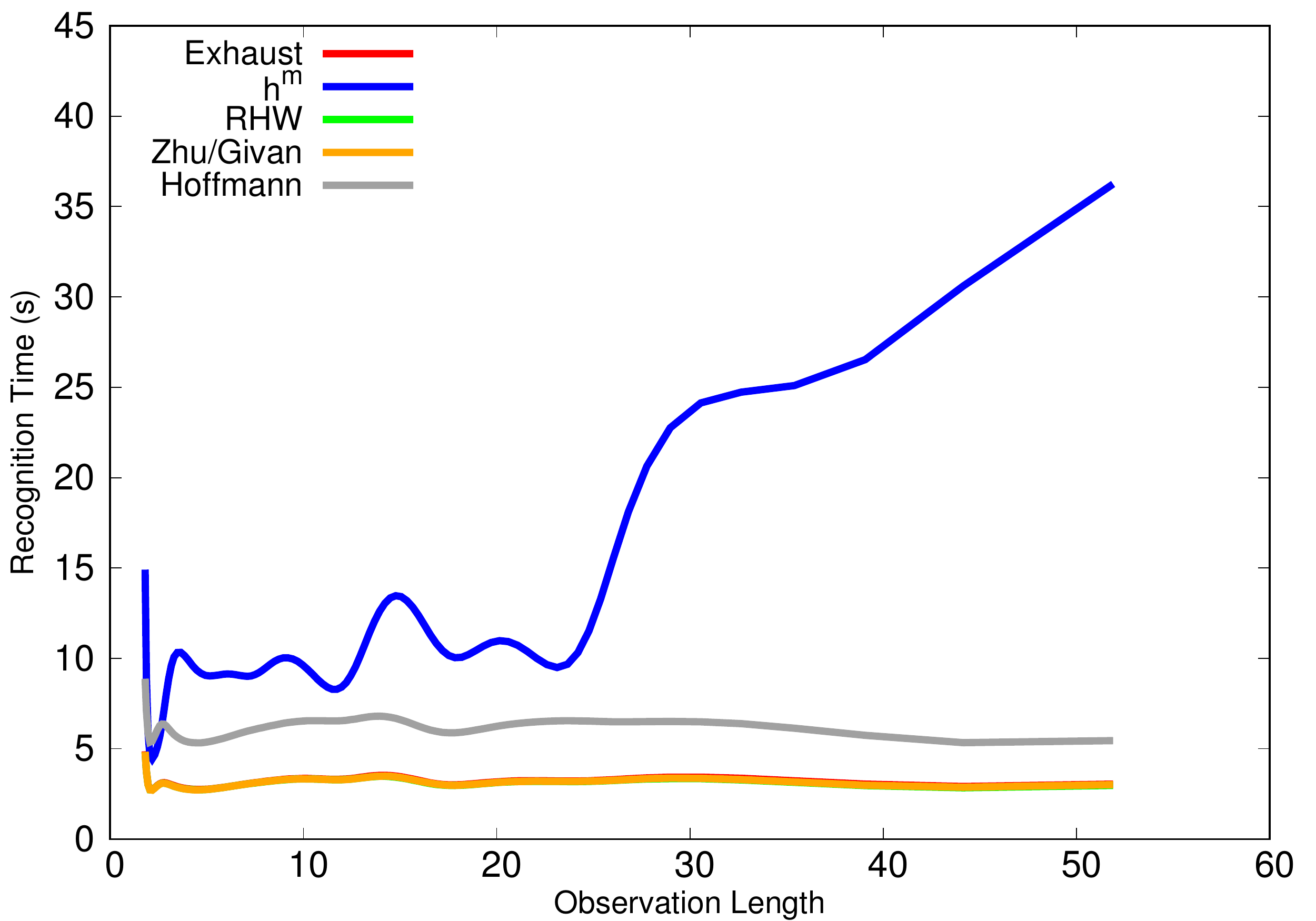}
\caption{Recognition time for $\mathit{h_{uniq}}$ with missing and full observations.}
\label{fig:time_uniq}
\end{figure}

Figures~\ref{fig:time_gc} and \ref{fig:time_uniq}, show how the recognition time varies with the growth of observation length for $\mathit{h_{gc}}$ and $\mathit{h_{uniq}}$, respectively. 
We can see that all algorithms provide a close to constant recognition time, except for $h^m$, in which we see the recognition time grows as the observation grows in length. Note that some curves are overlaid by others, causing them to not appear.

Note that the sequence of observations does not have a direct impact on the landmark extraction algorithms since they are not provided to the algorithms. 
However, longer observations generally translate to more complex problems, resulting in the increasing recognition time.

%Although $h^m$ shows the best performance in accuracy and %spread, it has the longest running time from all %algorithms. The second longest running time we see is in %Hoffmannn et al. solution. All other algorithms have %similar execution times, being faster than both Hoffmann et %al. and $h^m$.

%########################################################################
\subsection{Results: Noisy, Missing, and Full Observations}

In this section, we present and analyze the results obtained by experimenting the different recognition approaches in problems under noisy observations. We refer to noisy observations as a set of observed actions in which some of the actions are spurious actions. As mentioned before, for the datasets with noisy observations, we have 4 levels of observability, as follows: 25\%, 50\%, 75\%, and 100\%.

We can see the results for both recognition heuristics in Table~\ref{tab:results_noisy}. 
This table has the same format as the one presented in the previous section, for missing and full observations, the only difference is the number of columns, as now we have four observability levels instead of five.

\begin{table*}[ht!]
\centering
\fontsize{6.5}{11}\selectfont
\setlength\tabcolsep{3pt}
\begin{tabular}{rrrrr@{\hspace*{8mm}}rrr@{\hspace*{8mm}}rrr@{\hspace*{8mm}}rrr@{\hspace*{8mm}}rrl}
\toprule	
\hline

&
& \multicolumn{3}{c}{\bf 25\%}                                                             
& \multicolumn{3}{c}{\bf 50\%}                                                             
& \multicolumn{3}{c}{\bf 75\%}                                                             
& \multicolumn{3}{c}{\bf 100\%}                                                             \\ \hline

\textbf{Approach}
& $|\mathcal{L}|$
& \multicolumn{1}{r}{\bf Time} & \multicolumn{1}{r}{\bf Acc \%} & \multicolumn{1}{l}{\bf S in $\mathcal{G}$} 
& \multicolumn{1}{r}{\bf Time} & \multicolumn{1}{r}{\bf Acc \%} & \multicolumn{1}{l}{\bf S in $\mathcal{G}$} 
& \multicolumn{1}{r}{\bf Time} & \multicolumn{1}{r}{\bf Acc \%} & \multicolumn{1}{l}{\bf S in $\mathcal{G}$} 
& \multicolumn{1}{r}{\bf Time} & \multicolumn{1}{r}{\bf Acc \%} & \multicolumn{1}{l}{\bf S in $\mathcal{G}$} \\ \hline
      $\mathit{h_{gc}}$ (Exhaust $\theta = 0$) 
        & 29.8 & 4.823 & 47.5\% & 1.421
             & 5.787 & 73.5\% & 1.237
             & 5.148 & 87.3\% & 1.111
             & 5.924 & 95.7\% & 1.085
            \\
        $\mathit{h_{gc}}$ (Exhaust $\theta = 10$) 
        & 29.8 & 4.829 & 72.2\% & 3.149
             & 5.759 & 90.7\% & 2.832
             & 5.174 & 96.8\% & 2.356
             & 5.947 & 99.7\% & 2.171
            \\
        $\mathit{h_{gc}}$ ($h^m$ $\theta = 0$) 
        & 17.5 & 10.702 & 49.5\% & 1.526
             & 12.531 & 73.4\% & 1.272
             & 11.316 & 86.9\% & 1.121
             & 13.648 & 95.7\% & 1.125
            \\
        $\mathit{h_{gc}}$ ($h^m$ $\theta = 10$) 
        & 17.5 & 10.685 & 65.7\% & 2.693
             & 12.483 & 85.6\% & 2.326
             & 11.351 & 95.3\% & 1.944
             & 13.680 & 98.7\% & 1.752
            \\
        $\mathit{h_{gc}}$ (RHW $\theta = 0$) 
        & 19.7 & 4.778 & 48.8\% & 1.504
             & 5.726 & 72.9\% & 1.284
             & 5.126 & 86.8\% & 1.140
             & 5.895 & 95.1\% & 1.129
            \\
        $\mathit{h_{gc}}$ (RHW $\theta = 10$) 
        & 19.7 & 4.771 & 64.1\% & 2.535
             & 5.707 & 84.4\% & 2.244
             & 5.098 & 94.4\% & 1.926
             & 5.909 & 98.3\% & 1.692
            \\
        $\mathit{h_{gc}}$ (Zhu \& Givan $\theta = 0$) 
        & 16.8 & 4.814 & 48.3\% & 1.520
             & 5.737 & 73.4\% & 1.280
             & 5.116 & 86.9\% & 1.131
             & 5.921 & 94.7\% & 1.131
            \\
        $\mathit{h_{gc}}$ (Zhu \& Givan $\theta = 10$) 
        & 16.8 & 4.793 & 65.7\% & 2.655
             & 5.726 & 84.8\% & 2.330
             & 5.111 & 94.3\% & 1.966
             & 5.931 & 98.1\% & 1.700
            \\
        $\mathit{h_{gc}}$ (Hoffmann $\theta = 0$) 
        & 16.3 & 9.267 & 44.9\% & 1.465
             & 11.572 & 68.8\% & 1.282
             & 10.069 & 82.4\% & 1.185
             & 12.068 & 91.3\% & 1.153
            \\
        $\mathit{h_{gc}}$ (Hoffmann $\theta = 10$) 
        & 16.3 & 9.181 & 61.2\% & 2.462
             & 11.602 & 81.2\% & 2.256
             & 9.976 & 90.1\% & 1.965
             & 12.047 & 95.9\% & 1.748
            \\
        \hline
$\mathit{h_{uniq}}$ (Exhaust $\theta = 0$) 
        & 29.8 & 4.280 & 38.5\% & 1.098
             & 5.546 & 62.4\% & 1.069
             & 4.510 & 82.1\% & 1.050
             & 6.149 & 93.8\% & 1.050
            \\
        $\mathit{h_{uniq}}$ (Exhaust $\theta = 10$) 
        & 29.8 & 4.270 & 52.6\% & 1.768
             & 5.501 & 75.6\% & 1.612
             & 4.500 & 88.7\% & 1.369
             & 6.210 & 96.7\% & 1.337
            \\
        $\mathit{h_{uniq}}$ ($h^m$ $\theta = 0$) 
        & 17.5 & 9.050 & 39.1\% & 1.164
             & 11.096 & 62.7\% & 1.062
             & 9.407 & 82.3\% & 1.054
             & 12.711 & 94.2\% & 1.093
            \\
        $\mathit{h_{uniq}}$ ($h^m$ $\theta = 10$) 
        & 17.5 & 9.053 & 50.9\% & 1.734
             & 11.095 & 74.1\% & 1.598
             & 9.386 & 88.7\% & 1.354
             & 12.836 & 97.0\% & 1.293
            \\
        $\mathit{h_{uniq}}$ (RHW $\theta = 0$) 
        & 19.7 & 4.254 & 38.7\% & 1.152
             & 5.509 & 62.8\% & 1.064
             & 4.501 & 81.3\% & 1.059
             & 6.175 & 94.2\% & 1.095
            \\
        $\mathit{h_{uniq}}$ (RHW $\theta = 10$) 
        & 19.7 & 4.246 & 51.0\% & 1.747
             & 5.495 & 75.1\% & 1.609
             & 4.494 & 87.5\% & 1.354
             & 6.169 & 96.5\% & 1.303
            \\
        $\mathit{h_{uniq}}$ (Zhu \& Givan $\theta = 0$) 
        & 16.8 & 4.259 & 39.5\% & 1.172
             & 5.484 & 62.7\% & 1.063
             & 4.500 & 82.7\% & 1.062
             & 6.191 & 94.2\% & 1.101
            \\
        $\mathit{h_{uniq}}$ (Zhu \& Givan $\theta = 10$) 
        & 16.8 & 4.256 & 51.1\% & 1.756
             & 5.503 & 74.7\% & 1.606
             & 4.498 & 88.1\% & 1.364
             & 6.200 & 96.7\% & 1.301
            \\
        $\mathit{h_{uniq}}$ (Hoffmann $\theta = 0$) 
        & 16.3 & 7.903 & 36.4\% & 1.138
             & 11.022 & 59.7\% & 1.069
             & 8.500 & 77.0\% & 1.065
             & 12.859 & 88.3\% & 1.077
            \\
        $\mathit{h_{uniq}}$ (Hoffmann $\theta = 10$) 
        & 16.3 & 7.900 & 48.1\% & 1.742
             & 11.104 & 70.7\% & 1.612
             & 8.507 & 83.2\% & 1.419
             & 12.951 & 92.7\% & 1.346
            \\
        \hline
\bottomrule
\end{tabular}
\caption{Experiments and evaluations with missing, noisy and full observations.}
\label{tab:results_noisy}	
\end{table*}

We notice a drop in the accuracy metric by comparing the results in Tables~\ref{tab:results} and \ref{tab:results_noisy} and argue that it is an expected behavior, as the noise within the observations tends to mislead the recognition heuristics into recognizing the wrong goals as correct. 
Also, as expected, the recognition time is unaffected with relation to noiseless observations, with $h^m$ having the longest recognition times, followed by \textit{Hoffmann et al.} and the other algorithms. 
We can see that in noisy experiments, there is less difference between \textit{Hoffmann et al.} and $h^m$ recognition times.

\begin{figure}[h!]
\centering
\includegraphics[width=0.45\textwidth]{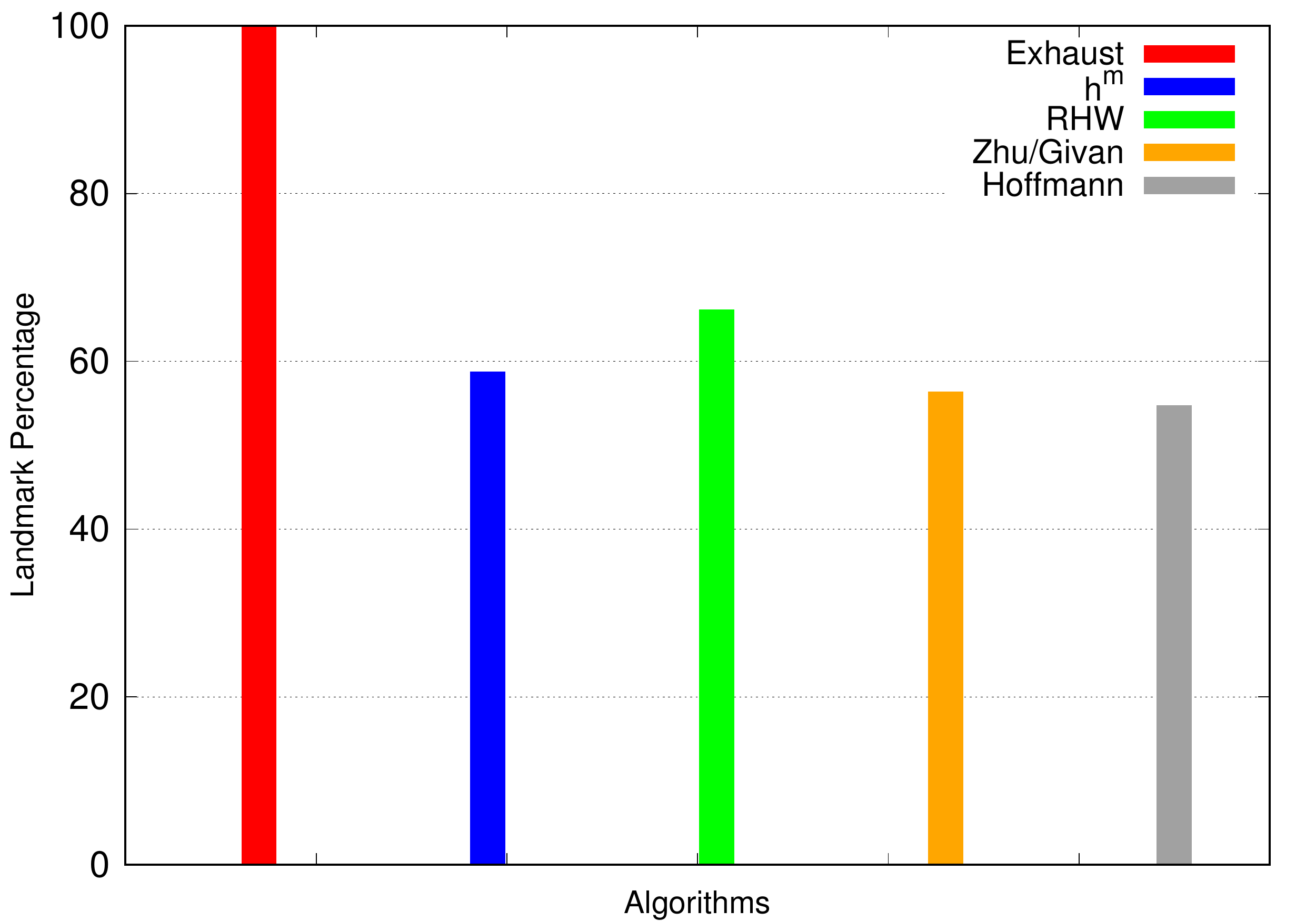}
\caption{Percentage of extracted landmarks by algorithm with missing, noisy and full observations.}
\label{fig:percentage_noisy}
\end{figure}

Figure~\ref{fig:percentage_noisy} shows the average percentage of landmarks extracted by each algorithm for the datasets with noisy observations. 
This metric has to be recalculated for noisy observations, as the goal recognition problems with noisy observations \emph{are different} from the ones without noise. 
We can see all algorithms, except for \textit{Exhaust}, managed to achieve a higher percentage of achieved landmarks in comparison to noiseless experiments, as the number of landmarks extracted by \textit{Exhaust} dropped. 
Yet, the algorithm ranking for the percentage of landmarks extracted remains similar to the noiseless experiments.

\begin{figure}[h!]
\centering
\includegraphics[width=0.45\textwidth]{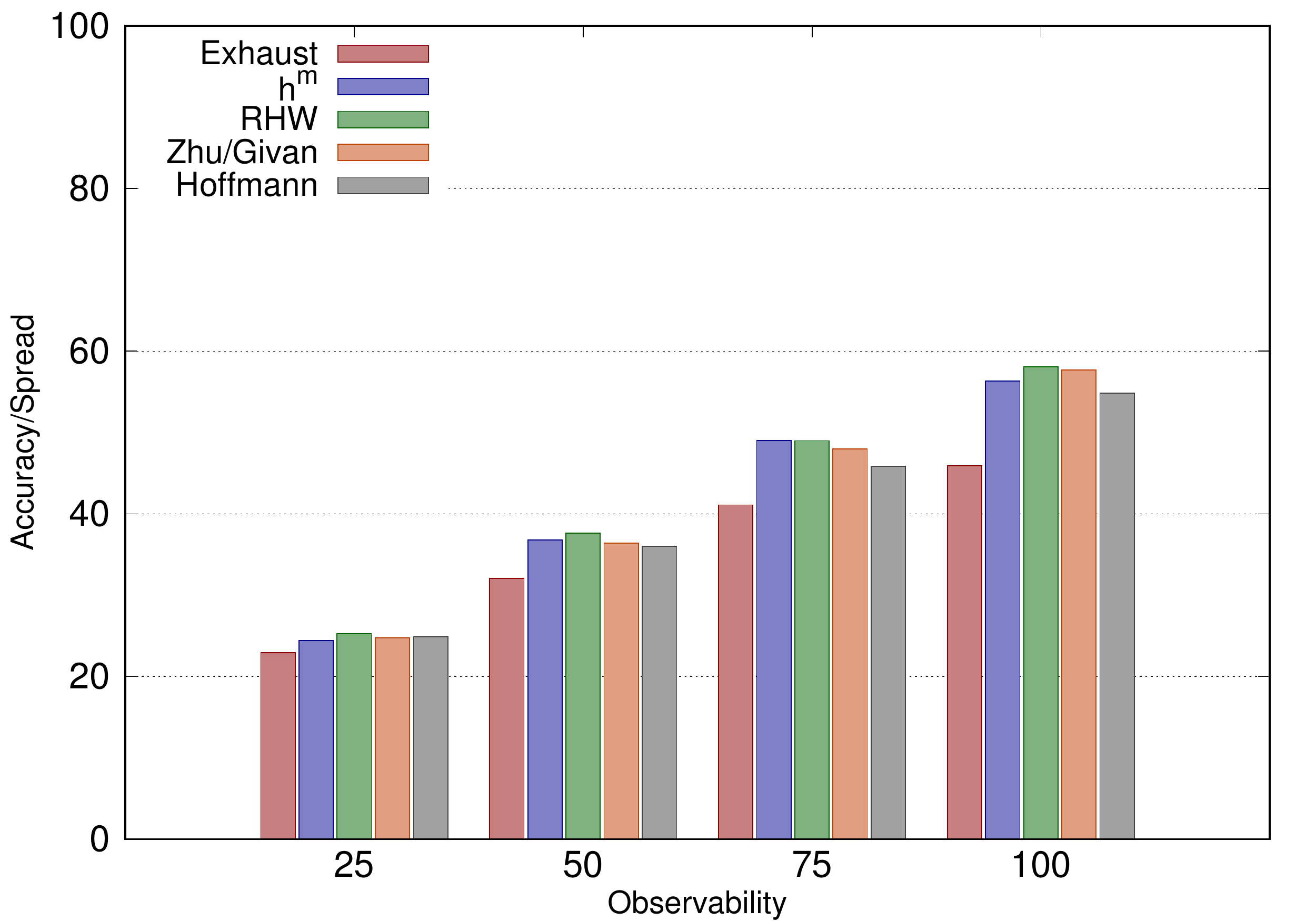}
\caption{Accuracy/Spread in $\mathcal{G}$ ratio for $\mathit{h_{gc}}$ with missing, noisy and full observations.}
\label{fig:accuracy_gc_noisy}
\end{figure}

\begin{figure}[h!]
\centering
\includegraphics[width=0.45\textwidth]{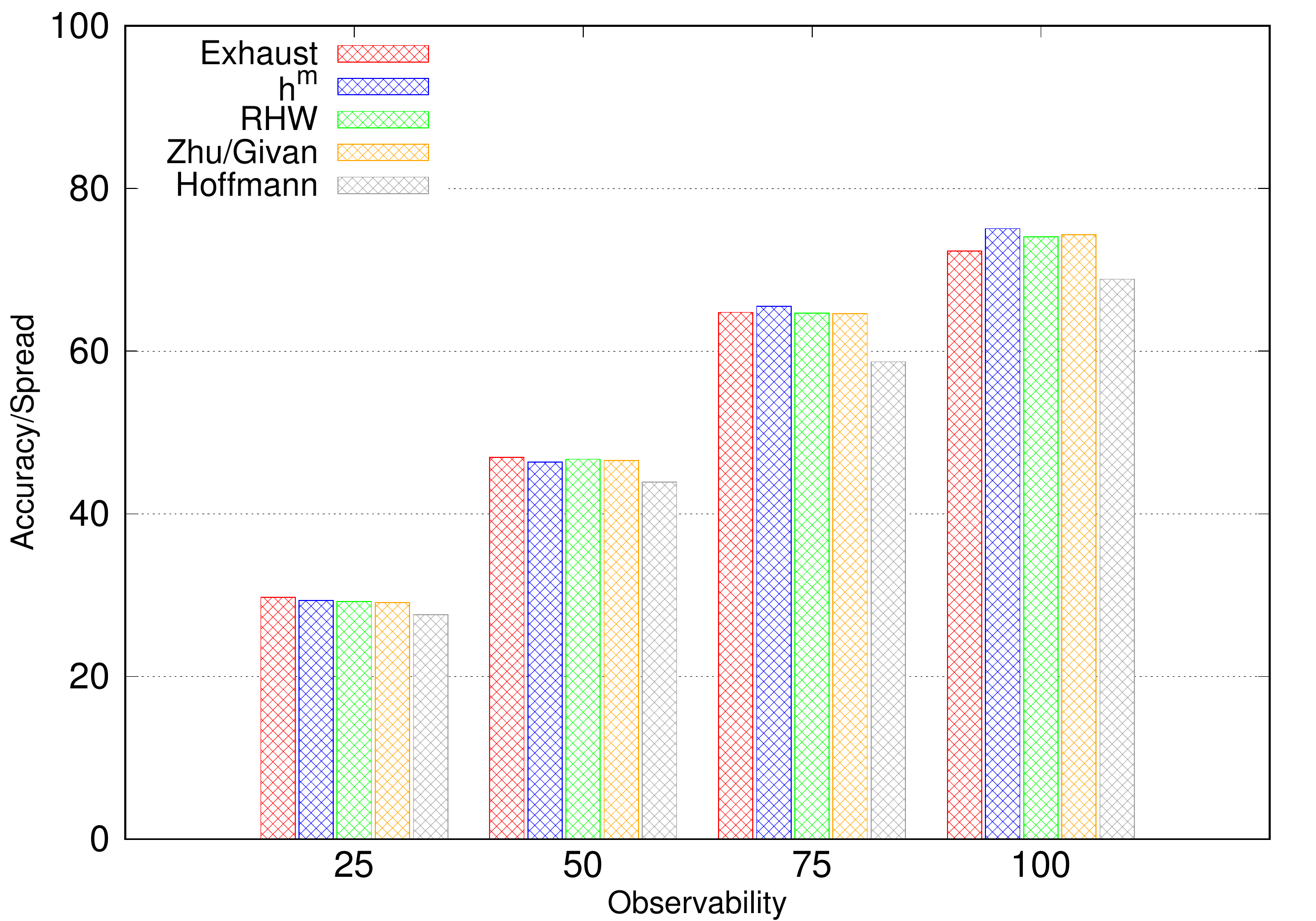}
\caption{Accuracy/Spread in $\mathcal{G}$ ratio for $\mathit{h_{uniq}}$ with missing, noisy and full observations.}
\label{fig:accuracy_uniq_noisy}
\end{figure}

Figures~\ref{fig:accuracy_gc_noisy} and \ref{fig:accuracy_uniq_noisy} show the Accuracy/Spread in $\mathit{G}$ ratio for each algorithm and observability degree for a threshold value of 10, for $\mathit{h_{gc}}$ and $\mathit{h_{uniq}}$ respectively.

As for the results using the $\mathit{h_{gc}}$ heuristic, we can see a different scenario when comparing to noiseless experiments. 
With noisy observations, the extraction algorithm that had the best overall performance in Accuracy/Spread in $\mathcal{G}$ was \textit{RHW}, which also extracted the most landmarks after \textit{Exhaust}. 
\textit{RHW} dominated the score for 25\% and 50\% observability levels, only being beaten by $h^m$ in 75\% and \textit{Zhu \& Givan} in 100\%.

With respect to the results of the $\mathit{h_{uniq}}$ heuristic results, we see the same behavior in noiseless experiments. 
Algorithms that extract a larger number of landmarks yielded better results in comparison to $\mathit{h_{gc}}$, as we can see from \textit{Exhaust} results in 25\% and 50\% observability levels, only being beaten by $h^m$ in 75\% and 100\%.

From these results, we can see how the presence of noise in observations really affects the recognition with different landmark extraction algorithms. 
When we work with noisy observations, the number of landmarks extracted seems have a stronger impact. 
This can be explained by the fact that having irrelevant actions within the observations makes so that having more landmarks may help the heuristic while comparing them against the relevant observations, as noisy actions are unlikely to coincide within the landmarks for the correct goal. 
% The recognition time is similar to noiseless experiments though.

\begin{figure}[h]
\centering
\includegraphics[width=0.45\textwidth]{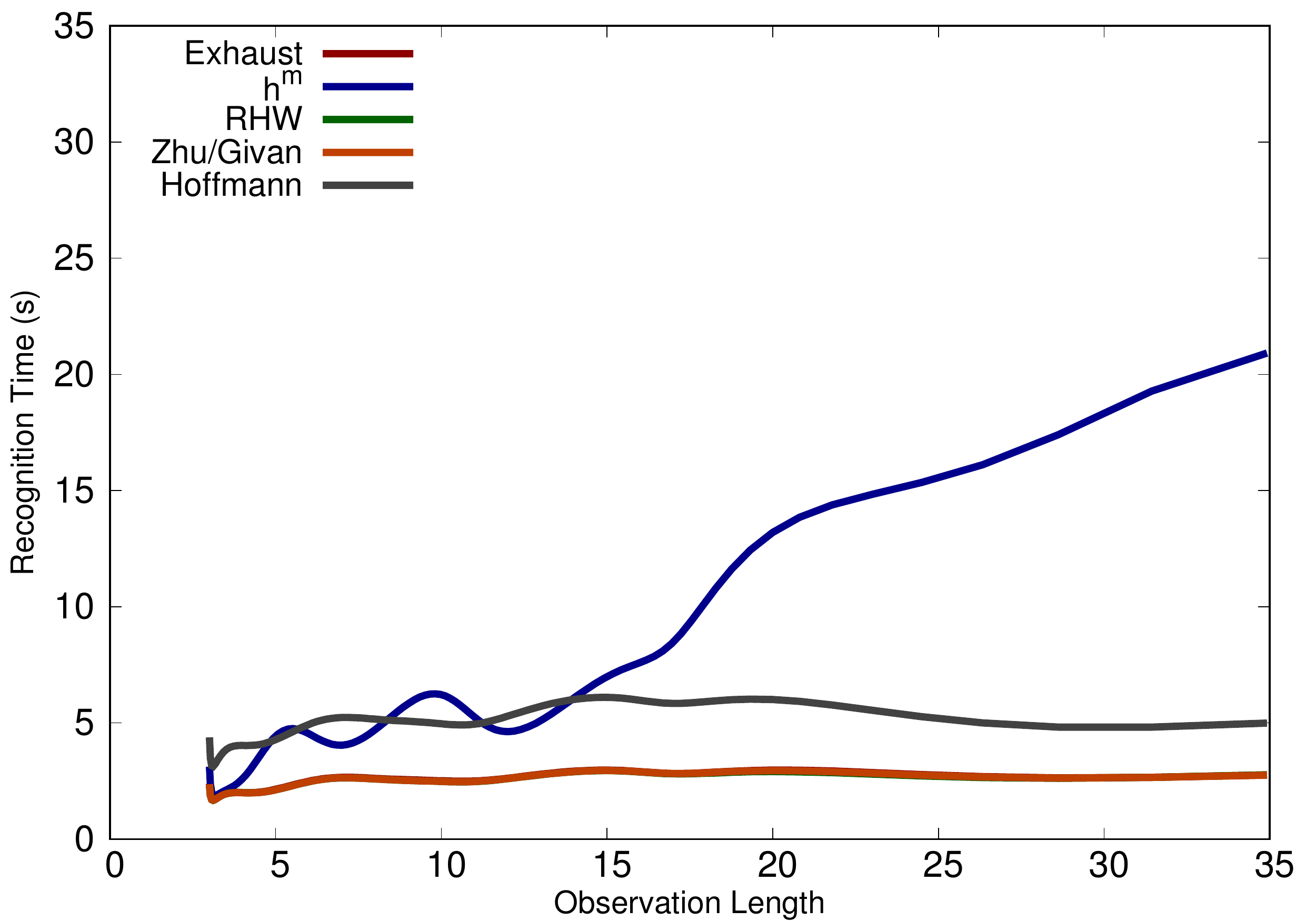}
\caption{Recognition time for $\mathit{h_{gc}}$ with missing, noisy and full observations.}
\label{fig:time_gc_noisy}
\end{figure}

\begin{figure}[h]
\centering
\includegraphics[width=0.45\textwidth]{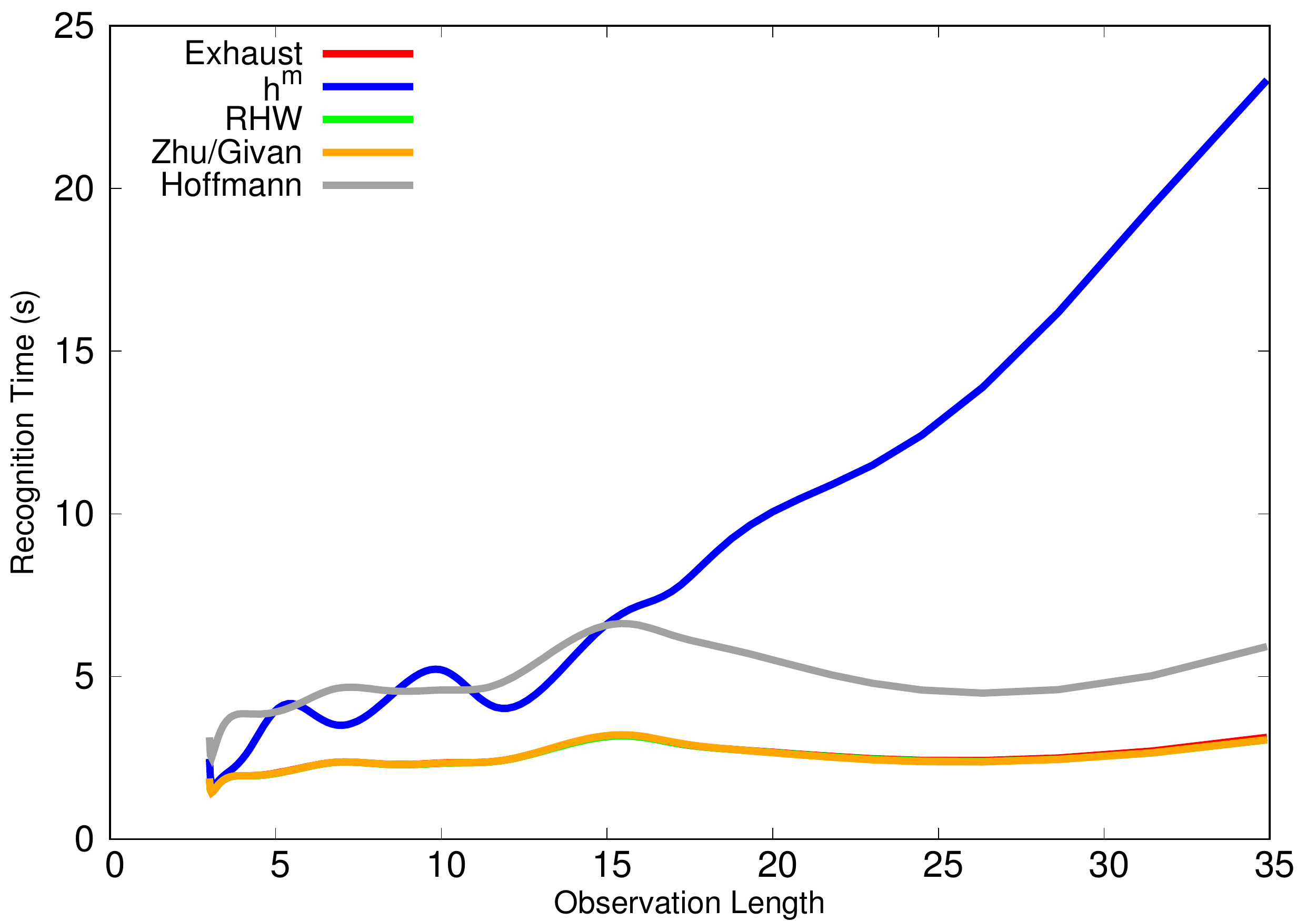}
\caption{Recognition time for $\mathit{h_{uniq}}$ with missing, noisy and full observations.}
\label{fig:time_uniq_noisy}
\end{figure}

Finally, in Figures~\ref{fig:time_gc_noisy} and~\ref{fig:time_uniq_noisy}, we can see the recognition time variation as observation length grows for $\mathit{h_{gc}}$ and $\mathit{h_{uniq}}$, respectively. 
A similar time marks can be seen without noisy observations, with $h^m$'s running time growing with observation length, while the other algorithms remain almost constant, with minor differences. We also see the same curve overlay effect that causes some curves to not appear.

%----------------------------------------------------------------------------
\section{Conclusions}\label{section:conclusions}

We have presented an extensive empirical evaluation of how different landmark extraction algorithms affect the performance of landmark-based goal recognition approaches. 
After analyzing the results in the experiments, we conclude that the number of extracted landmarks does not tell us all about the quality or utility of a landmark when using it in landmark-based goal recognition. 
We can see from the results that having more landmarks is not necessarily more important than having informative landmarks.

As future work, we intend to perform a more qualitative analysis of the landmark extraction algorithms, analyzing not only the amount of extracted landmarks, but also the information level of the landmarks themselves. 
This ought to provide even more answers on what kind of extraction algorithm is best suited for landmark-based goal recognition, and consequently enabling us to fine-tune solutions to maximize the effectiveness of the goal recognition process. 
Finally, we aim to conduct a similar extensive empirical evaluation by using some of the landmark extraction algorithms over the landmark-based approaches under incomplete domain information~\cite{RamonMeneguzzi_AAAI2018,PereiraPM19}.

%----------------------------------------------------------------------------
\bibliographystyle{aaai}
\bibliography{pair2020-goal_recognition}

\end{document}